\documentclass[11pt,a4paper]{article}
\usepackage[hyperref]{acl2018}
\usepackage{times}
\usepackage{latexsym}
\usepackage{amsmath}
\DeclareMathOperator*{\argmin}{arg\,min} 
\DeclareMathOperator*{\argmax}{arg\,max} 
\usepackage[linesnumbered,ruled]{algorithm2e}
\usepackage{booktabs} 
\usepackage{latexsym}
\usepackage{fixltx2e}
\usepackage{subcaption}
\usepackage{multirow}
\usepackage{caption}
\usepackage{graphicx}
\usepackage{url}

\aclfinalcopy 
\def\aclpaperid{***} 

\title{Multi-Task Active Learning for Neural Semantic Role Labeling on Low Resource Conversational Corpus}

\author{Fariz Ikhwantri$^1$ \qquad Samuel Louvan$^{1,2}$ \qquad Kemal Kurniawan$^1$ \qquad Bagas Abisena$^1$\\ \textbf{Valdi Rachman$^{3}$} \qquad
\textbf{Alfan Farizki Wicaksono$^3$ \qquad Rahmad Mahendra$^3$}\\
  $^1$Kata.ai, Jakarta, Indonesia \\
  $^2$Fondazione Bruno Kessler/University of Trento, Trento, Italy \\
  $^3$Universitas Indonesia, Depok, Indonesia \\
  {\tt \{fariz,kemal,bagas\}@kata.ai, slouvan@fbk.eu}\\
  {\tt valdi.rachman@gmail.com}\\
  {\tt \{alfan,rahmad.mahendra\}@cs.ui.ac.id} 
  }

\date{}

\begin{document}
\maketitle
\begin{abstract}
Most Semantic Role Labeling (SRL) approaches are supervised methods which require a significant amount of annotated corpus, and the annotation requires linguistic expertise. In this paper, we propose a Multi-Task Active Learning framework for Semantic Role Labeling with Entity Recognition (ER) as the auxiliary task to alleviate the need for extensive data and use additional information from ER to help SRL. We evaluate our approach on Indonesian conversational dataset. Our experiments show that multi-task active learning can outperform single-task active learning method and standard multi-task learning. According to our results, active learning is more efficient by using 12\% less of training data compared to passive learning in both single-task and multi-task setting. We also introduce a new dataset for SRL in Indonesian conversational domain to encourage further research in this area\footnote{request to \href{mailto:research@kata.ai}{research@kata.ai}}.
\end{abstract}



\section{Introduction}

Semantic Role Labeling (SRL) extracts predicate-argument structures from sentences \cite{Jurafsky2006SpeechAL}. It tries to recover information beyond syntax. In particular, information that can answer the question about “who did what to whom, when, why and so on” \citep{Johansson:2008:DSR:1613715.1613726,CHOI10.73}.

There have been many proposed SRL techniques, and the high performing models are mostly supervised \cite{Akbik2016KSRLIL, Punyakanok2004SemanticRL}. As they are supervised methods, the models are trained on a relatively large annotated corpus. Building such corpus is expensive as it is laborious, time-consuming, and usually requires expertise in linguistics. For example, PropBank annotation guideline by \citet{CHOI10.73} is around 90 pages so it can be a steep learning curve even for annotators with a linguistic background. This difficulty makes reproducibility hard for creating annotated data especially in low resource language or different domain of data. Several approaches have been proposed to reduce the effort of annotation. \citet{He2015QuestionAnswerDS} introduced a Question Answering-driven approach by casting a predicate as a question and its thematic role as an answer in the system. \citet{Wang2017ActiveLF} used active learning using semantic embedding. \citet{Akbik2017} utilized Annotation Projection with hybrid crowd-sourcing to route between hard instances for linguistic experts and easy instances for non-expert crowds.

Active Learning is the most common method to reduce annotation by using a model to minimize the amount of data to be annotated while maximizing its performance. In this paper, we propose to combine active learning with multi-task learning applied to Semantic Role Labeling by using a related linguistic task as an auxiliary task in an end-to-end role labeling. Our motivation to use a multi-task method is in the same spirit as \cite{Gormley2014} where they employed related syntactic tasks to improve SRL in low-resource languages as multi-task learning. Instead, we used Entity Recognition (ER) as the auxiliary task because we think ER is semantically related with SRL in some ways. For example, given a sentence: \emph{Andy gives a book to John}, in SRL context, \emph{Andy} and \emph{John} are labeled as AGENT and PATIENT or BENEFACTOR respectively, but in ER context, they are labeled as PERSON. Hence, although the labels are different, we hypothesize that there is some useful information from ER that can be leveraged to improve overall SRL performance.

Our contribution in this paper consists of two parts. First, we propose to train multi-task active learning with Semantic Role Labeling as the primary task and Entity Recognition as the auxiliary task. Second, we introduce a new dataset and annotation tags for Semantic Role Labeling from conversational chat logs between a bot and human users. While many of the previous work studied SRL on large scale English datasets in news domain, our research aims to explore SRL in Indonesian conversational language, which is still under-resourced.

\section{Related Work}

\subparagraph{Active learning} (AL) \cite{Settles:10} is a method to improve the performance of a learner by iteratively asking a new set of hypotheses to be labeled by human experts. A well-known method is Pool-Based AL, which selects the hypotheses predicted from a pool of unlabeled data \cite{Lewis1994ASA}. The most informative instance from hypotheses is selected and added into labeled data. The informativeness of an instance is measured by its uncertainty, which is inversely proportional to the learner's confidence of its prediction for that instance. In other words, the most informative instance is the one which the model is least confident with. 

There are two well-studied methods of sequence labeling with active learning. The first one is maximum entropy: given an input sentence $x$, the probability of word $x_t$ having tag $y_t$ is given by 

\begin{equation} \label{eq:softmax}
p_\theta(y_t|x_t) = \frac{exp(~a^{y_t}_t(x_t|\theta))}{\sum_{j=1}^{K}exp(~a_t^j(x_t|\theta))}
\end{equation}
Where $\theta$ denotes a model parameters and $K$ is the number of tags. Uncertainty in maximum entropy can be defined using Token Entropy (TE) as described in \cite{Settles:2008:AAL:1613715.1613855,Marcheggiani2014AnEC}.
\begin{equation} \label{eq:token-log-entropy}
\phi_{t}^{\mathrm{TE}} = - \sum_{j \in K}p(y_t=j|x_t)\log p(y_t=j|x_t)
\end{equation} 
\begin{equation} \label{eq:instance-token-entropy}
x_{\mathrm{TE}}=\argmax_{x} \sum_{t=1}^T -\phi_t^{\mathrm{TE}}
\end{equation}

From token level entropy ($\mathrm{TE}$) in \eqref{eq:token-log-entropy}, we used a simple aggregation such as summation to select an instance. So that instance $x$ is selected by Equation~\eqref{eq:instance-token-entropy} as least confident sample, where $\sum_{t=1}^T (.)$ is a summation term for greedy aggregation of sentence level entropy.

Another well-studied sequence labeling method with active learning is Conditional Random Fields (CRFs) by \citet{Lafferty2001ConditionalRF}, where the probability of a sequence label $\mathbf{y}=\{y_1,y_2,..,y_T\}$ given a sequence of observed vectors $\mathbf{x} =\{x_1,x_2,..,x_T\}$ and a joint log-likelihood function of unary and transition parameter $\psi(y_{t-1}, y_t, x_t)$ is defined as
\begin{equation} \label{eq:crf}
p_\psi(y|x) = \frac{\prod_{t=1}^T \psi(y_{t-1}, y_t, x_t)}{\sum_{y \in Y} \prod_{t=1}^T \psi(y_{t-1}, y_t, x_t)} 
\end{equation}

Uncertainty in conditional random fields can be obtained by Viterbi decoding by selecting instance with maximum $p(y|x)$ from a pool of unlabeled instances as defined below.
\begin{equation}
x_{\mathrm{VE}} = \argmin_{x}p_{\psi}(y^{\star}|x)
\label{eq:min-viterbi}
\end{equation}
where $p(y^{\star}|.)$ is a probability assigned by Viterbi inference algorithm \cite{Marcheggiani2014AnEC}. 

\subparagraph{Multi-Task Learning}

Instead of training one task per model independently, one can use related labels to optimize multiple tasks in a learning process jointly. This method is commonly known as Multi-Task learning (MTL) or as Parallel Transfer Learning \cite{Caruana1997MultitaskL}. Our motivation to use multi-task learning is to leverage "easier" annotation than Semantic Roles to regularize model by using related tasks. Previous work on Multi-Task learning on Semantic Role Labeling by \citet{Collobert:2011:NLP:1953048.2078186} did not report any significant improvement for SRL task. A recent work \cite{DBLP:journals/corr/abs-1711-00768} used SRL as the auxiliary task with Opinion Role Labeling as the main task. 

\subparagraph{Multi-Task Active Learning} 
Previous work on multi-task active learning (MT-AL) \cite{reichart-EtAl:2008:ACLMain} was focused on formulating a method to keep the performance across a set of tasks instead of a single task. In multi-task active learning scenario, optimizing a set of task classifiers can be regarded as a meta-protocol by combining each task query strategy into a single query method. In one-sided task query scenario settings, one selected task classifier uncertainty strategy is used to query unlabeled samples. In multiple task scenario, the uncertainty of an instance is the aggregate of classifier’s uncertainties for all tasks.

\begin{figure}[t]
\includegraphics[scale=0.3]{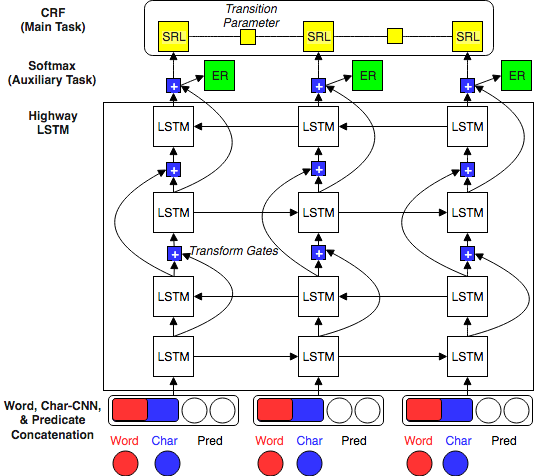}
\caption{Model Overview. Four layers Highway LSTM. SRL task used Conditional Random Fields (CRF) for sequence labeling output.}
\label{fig:model}
\end{figure}

\section{Proposed Method}
In this section, we explain on how we incorporated both the AL and MTL in our neural network architecture. We used the state-of-the-art SRL model from \citet{he2017deep} as our base model as shown in Figure \ref{fig:model}.

Our model is a modification of \citeauthor{he2017deep}'s work. Our first adjustment is to use  CRF as the last layer instead of softmax because of its notable superiority found by \citet{Reimers2017ReportingSD} for both role labeling and entity recognition. In this scenario, we used CRF layer for the primary task (SRL) \cite{zhou-xu:2015:ACL-IJCNLP} and softmax layer for the auxiliary task. The auxiliary task acts as a regularization method \citep{Caruana1997MultitaskL}. Second, we used character embedding with Convolutional Neural Networks as Characters Encoder \cite{Ma2016EndtoendSL}, to handle out-of-vocabulary problem caused by misspelled words, slangs, and abbreviations common in informal chatting, as well as word embedding and predicate indicator feature embedding as the input features for a Highway LSTM.

In multi-task learning configuration, we used parameter sharing in embedding and sequence encoder layers except for the outermost module which is used for prediction for each specific task. We optimized the parameters jointly by minimizing the sum loss of $L(y_{s}, y_{e}| x, \theta, \psi) =L(\hat{y_{s}}, y_{s}|x, \theta) + L(\hat{y_{e}}, y_{e}| x, \psi)$, where the first part of the equation is the SRL loss and the second part is the entity loss. SRL loss is computed by joint log-likelihood of emissions with transition parameters in CRF from Equation~\ref{eq:crf} and entity loss is computed using standard cross-entropy loss from softmax output in Equation~\ref{eq:softmax}.

\subparagraph{Multi-Task Active Learning} 
 In multiple task scenario, we used the rank combination by \citet{reichart-EtAl:2008:ACLMain} that combines each task query strategy into an overall $\mathrm{rank}(x^{i}) = \mathrm{rank}(x_{\mathrm{VE}}^{i}) + \mathrm{rank}(x_{\mathrm{TE}}^{i})$. Note that in both training one-sided and combined rank multi-task active learning, we returned all task gold labels to be trained in multi-task models. 

As a multi-task active learning baseline, instead of one-sided AL which queries a pre-determined task for all-iteration, we used random task selection to draw which task to use as the query strategy in the $i$-th iteration. Random task selection is implemented using random multinomial sampling. The selected task is used for the query instances using standard uncertainty sampling.

\section{Dataset \& Experiment}

\subsection{Dataset}
This research presents the dataset of human users conversation with virtual friends bot\footnote{https://kata.ai/case-studies/jemma}. The annotated messages are user inquiries or responses to the bot. Private information in the original data such as name, email, and address will be anonymized. Three annotators with a linguistic background performed the annotation process. In this work, we used a set of semantic roles adapted for informal, conversational language. Table \ref{tab:semantic_roles} shows some examples of the semantic roles. The dataset consists of 6057 unique sentences which contain predicates.

\begin{table}
\small
	\caption{Semantic Roles dataset for conversational language statistics and examples}
	\label{tab:semantic_roles}
	\begin{tabular}{@{}llp{3.3cm}@{}}
		\toprule
		Semantic Roles		& Count &Example\\
		\midrule
		AGENT (A)				& 2843 & \emph{I} brought you a present\\
		PATIENT	(PS)			& 3040 & I brought you \emph{a present}\\
		BENEFACTOR (BN)			& 293 & I brought \emph{you} a present\\
		GREET (G)				& 572 & Hi \emph{Andy}! \\ & & I brought you a present\\
		LOCATION (L)			& 183 & I can eat at \emph{home} today \\
		TIME (T)			& 399 & I can eat at home \emph{today} \\
		\bottomrule
	\end{tabular}
\end{table}
The semantic roles used are a subset of PropBank \cite{Palmer2005ThePB}. Also, we added a new role, GREET. In our collected data, Indonesian people tend to call the name of the person they are talking to. Because such case frequently co-occurs with another role, we felt the need to differentiate this previously mentioned entity as a new role. For example, in the following sentence: ”\emph{Hi Andy! I brought you a present}” can help refers "\emph{you}" role as PATIENT to "\emph{Andi}" role as GREET instead of left unassigned.

In our second task, which is Entity Recognition (ER), we annotated the same sentence after the SRL annotation. We used common labels such as \texttt{PERSON}, \texttt{LOCATION}, \texttt{ORGANIZATION}, and \texttt{MISC} as our entity tags. Different from Named Entity Recognition (NER), ER also tag nominal objects such as \emph{"I"}, \emph{"you"} and referential locations like \emph{"di sana (over there)"}. While this tagging might raise a question whether there are overlapping tags with SRL, we argue that entity labels are less ambiguous compared to role arguments which are dependent on the predicate. An example of this case can be seen in Table~\ref{tab:semantic_roles}, where both of \emph{I} and \emph{you} are tagged as PERSON whereas the roles are varied. In this task, we used semi-automatic annotation tools using \texttt{brat} \cite{stenetorp2012}. These annotation were checked and fixed by four people and one linguistic expert. 

\subsection{Experiment Scenario}

The purpose of the experiment is to understand whether multi-task learning and active learning help to improve SRL model performance compared to the baseline model (SRL with no AL scenario). In this section, we focus on several experiment scenarios: single-task SRL, single-task SRL with AL, MTL, and MTL with AL.

\subparagraph{Model Architecture} 
Our model architecture consists of word embedding, character 5-gram encoder using CNN and predicate embedding as inputs, with 50, 50, and 100 dimension respectively. These inputs are concatenated into a 200-dimensional vector which then fed into two-layer Highway LSTM with 300 hidden units.

\subparagraph{Initialization}
The word embedding were initialized with unsupervised pre-trained values obtained from training word2vec \cite{mikolov2013efficient} on the dataset. Word tokens were lowercased, while characters were not.

\subparagraph{Training Configurations} 
For training configurations, we trained for 10 epochs using AdaDelta \cite{Zeiler2012ADADELTAAA} with $\rho=0.95$ and $\epsilon=1.\mathrm{e}{-6}$. We also employed early stopping with patience set to 3. We split our data using 80\% training, 10\% validation, and 10\% test for the fully supervised scenario. For the active learning scenario, we further split the training data into labeled and unlabeled data. We used two kinds of split, 50:50 and 85:15. For the 50:50 scenario, we queried 100 sentences for each epoch. For the 85:15 scenario, we used a smaller query of 10 sentences in an epoch to keep the number of queries less than the number of available fully supervised training data in 10 epochs. This number of queried sentences was obtained by tuning on the validation set.

As for the AL query method, in the single-task SRL, we used random and uncertainty sampling query. SRL with 100\% training data and SRL with random query serve as baseline strategies. In the MTL SRL, we employed random task and ranking.

\section{Results \& Analysis}
\begin{table}[t]
\small
\begin{tabular}{@{} p{1.1cm} c p{0.85cm} | p{0.85cm}  p{0.85cm}  p{0.9cm} @{}}
  \toprule
  \multicolumn{3}{c}{Scenario}  & \multicolumn{3}{c}{Metric}    \\
  	\midrule
  Task & Active  & Data (\%) & P & R &  F1 \\
  	\midrule
  SRL &  - & 100 & 75.12 & 75.49 & 75.30  \\
  \hline
  SRL &  Random & 50 & 75.50 & 74.01 & 74.75  \\
  SRL &  Random & 85 & 78.83 &	71.91 &	75.21 \\
  SRL &  Uncertain & 50 & 76.67 & 74.01 & 75.32\\
  SRL &	Uncertain & 85 & 78.35 & 75.25 & 76.77\\
  \midrule
  SRL+ER & -	& 100 & 76.88	& 74.50	& 75.67\\
  \midrule
  SRL+ER & RandTask & 50 & 77.31 & 71.28 & 74.18\\
  SRL+ER & RandTask & 85 & 76.59 & 74.50 & 75.53\\
  SRL+ER & Ranking & 50 & 78.94 & 71.90 & 75.25\\
  SRL+ER & Ranking & 85 & 78.18 & 75.87 & \textbf{77.01}\\
\end{tabular}
\caption{Experiment results, Scenario Active means the query strategy used to sort instance informativeness, RandTask = Random Task Selection, Data scenario are initial percentage of labeled data, 50\% means the 50:50 split, 85\% means 85:15 split, and 100\% means use all training data. P (Precision), R (Recall), F1 (F1 Score)}
\label{tab:experiment}
\end{table}
\begin{figure}
\includegraphics[scale=0.39]{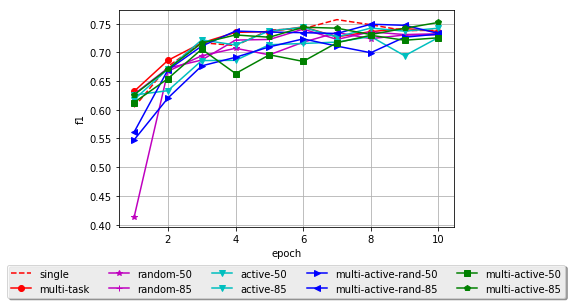}
\centering
\caption{Comparison of experiment scenarios in validation set. Multi-Task AL using Ranking Combination with initial 85\% labeled training data achieve best F1 score}
\label{fig:plot-compare}
\end{figure}
We experimented with a low-resource conversational language by varying the task scenario, active learning query strategy, and outset percentage of data seed from training data. We report our results using Precision (P), Recall (R), and the F1 score (F1) computed by exact matching of gold and predicted role spans. The report can be seen in Table \ref{tab:experiment}.

Our baseline multi-task (SRL+ER with no AL scenario) learning model in this experiment has a higher precision compared to the single-task (SRL) model. From the initial 85\% of labeled training data scenario, our model in total requested 87\% of the training data in 10 epochs. In this scenario, our proposed method for multi-task active learning using ranking combination can outperform the single-task active learning models. Figure \ref{fig:plot-compare} presents the F1 score learning curve for each model.

\begin{table}
\fontsize{7.0}{7.2}\selectfont
    \centering
    \begin{tabular}{lccc|ccc}
        \toprule
        Label & \multicolumn{3}{c}{dev} & \multicolumn{3}{c}{test} \\
        \cmidrule{2-4} \cmidrule{5-7}
        & P & R & F1 & P & R & F1\\
        \midrule
AGENT      & 87.03       & 83.43    & 85.196 & 86.68        & 85.85     & 86.26 \\
PATIENT     & 72.80         & 69.64    & 71.19 & 74.00        & 70.76      & 72.34 \\
BENEFACTOR & 60.53       & 76.67    & 67.65 & 38.10        & 42.11     & 40.00     \\
GREET      & 75.81      & 65.28    & 70.15 & 83.05        & 76.56     & 79.66 \\
LOCATION     & 50.00           & 34.62    & 40.91 & 60.00            & 65.22     & 62.50   \\
TIME      & 66.67       & 61.11    & 63.76 & 72.73        & 65.31     & 68.82 \\
        \bottomrule
    \end{tabular}
    \caption{Detailed scores of Multi-Task Active Learning performance with 85\% initial data. P (Precision), R (Recall), F1 (F1 Score)}
    \label{per-label}
\end{table}

\begin{figure}[t]
\includegraphics[scale=0.30]{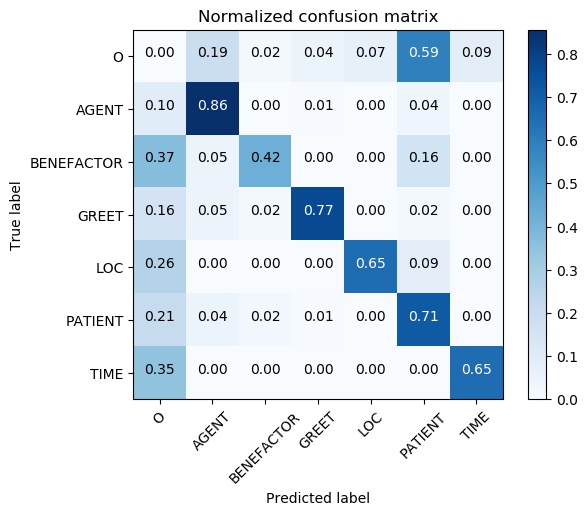}
\caption{Confusion matrix for Multi-Task Active Learning Model using 85\% initial labeled data.}
\label{fig:confusion-mat}
\end{figure}

\subparagraph{Significance test}
We performed two tails significance test (t-test) by using 5-fold cross validation from the training and the test parts of the corpus. The multi-task learning model is better compared to the single-task learning one $(p<0.05)$. However, the single-task and the multi-task learning scenario are not significantly better than both multi-task active learning from 85\% and 50\% training data scenario, since the $p$-value between model pairs are greater than 0.05. Therefore, accepting the null hypothesis indicate that performances between multi-task active learning with 50\%/85\% initial data and multi-task or single-task with full dataset are comparable.

We draw a confusion matrix of the multi-task active learning model with 85\% initial training data in Figure \ref{fig:confusion-mat} to analyze our model performance. We observe several common errors made by the model. The largest mistakes from the matrix are PATIENT false positive. The model incorrectly detected 59\% of non-roles as PATIENT. Another prominent error is 21\% false negative of total gold roles. The model primarily failed to tag 37\% of gold BENEFACTOR and 35\% of gold TIME. Quite different from the English SRL, we found that labels confusion happens less frequently than other error types. Based on this percentage, we investigated the error by drawing samples. In general, we broke down the incorrect predictions into several types of error.

\begin{table*}[ht]
\begin{minipage}{\textwidth}
\parbox{\linewidth}{
  \small
  \centering
  \begin{tabular}{llcc|llcc@{}}
	\multicolumn{4}{l}{Predicate : \emph{menjawab} (EN: reply)} & \multicolumn{4}{l}{Predicate : \emph{di donlot} (EN: download)}\\
\toprule
		token	& vocab & gold & predicted & token	& vocab & gold & predicted \\
\midrule
Yang  &     Yang   &    O      &   O & Jemma &  Jemma  &B-A  & B-A\\
menjawab &  UNK  &    O    &   O & udah  &  udah &  O        & O\\
ini   &     ini    &    B-PS & O & di  &    di  &   O      & O\\
komputer &  komputer &  I-PS & O & donlot & UNK &  O      & O\\
kan   &     kan    &    O   &      O  & get   &  UNK  & B-PS & O\\
?     &     ?      &    O   &      O & rick   &  UNK  &I-PS & O\\
      &            &        &        & nya    &  nya  & I-PS & O\\
  \end{tabular}
  \caption{Undetected Roles examples. Left translation: "This is a bot replying, right ?". Right translation: "Jemma, have you downloaded get rick?"}
  \label{add-roles}
}
\parbox{.45\linewidth}{
\small
\centering
\begin{tabular}{llcc@{}}
\multicolumn{4}{l}{Predicate: \emph{ada} (EN: exists)}\\
\toprule
		token	& vocab & gold & predicted \\
\midrule
jem  &    jem   &   B-G  & B-G\\
ada  &    ada   &   O    &   O\\
info &    info  &   B-PS & B-PS\\
makanan & makanan & I-PS & O\\
gak  &    gak   &   O    &   O\\
?    &    ?     &   O    &   O\\
.    &    .     &   O    &   O\\
&&&\\

\multicolumn{4}{l}{Predicate: \emph{tny} (EN : ask)}\\
\toprule
		token	& vocab & gold & predicted \\
\midrule
Aku  &    Aku   &   B-A  &  B-A \\
mau  &    mau   &   O    &   O\\
tny  &    UNK &     O    &   O\\
sahabar & UNK &   B-PS & B-PS\\
virtual & virtual & I-PS & O\\
itu   &   itu   &   I-PS & O\\
mksd  &   mksd  &   O    &     O\\
a     &   a     &   O    &     O\\
gimana &  gimana &  O    &     O\\
?    &    ?     &   O    &     O\\
.    &    .     &   O    &     O\\
        
\end{tabular}
\caption{Boundary error examples. Top translation: "Jem, do you have any food related info?". Bottom translation: "I want to ask what is a virtual friend meaning?"}
\label{fix-boundary}
}
\hfill{}
\parbox{.45\linewidth}{
\small
\centering
\begin{tabular}{llcc@{}}

\multicolumn{4}{l}{Predicate: \emph{genit} (EN : flirt)}\\

\toprule
true    &      vocab  &  gold      &   pred\\
\midrule
Jemma   &      Jemma  & B-P &   B-A\\
jangan  &      jangan & O    &   O\\
genit   &      UNK  &  O     &   O\\
sama    &      sama &  O     &   O\\
NAME  & UNK  &  B-BN & B-PS\\
;       & UNK  &  O        &   O\\
(       & (    &       O   &   O\\
.       & .    &       O   &   O\\
&&&\\

\multicolumn{4}{l}{Predicate: \emph{lihat} (EN : see)}\\
\toprule
true   & vocab&  gold&          predicted\\
\midrule
Aku  &  Aku  &  O   &  B-A \\
kesal&  UNK  &  O   &  O \\
lihat&  lihat&  O   &  O \\
kamu &  kamu &  B-G &  B-BN \\
dek  &  UNK & B-PS &   O \\
.   &   .   & I-PS &   O \\
\end{tabular}
\caption{Role confusion examples. Top translation: "Jemma do not flirt with NAME (person name is censored due to privacy)". Bottom translation: "I am annoyed to see you boy\footnote{In the original language, the word is gender neutral.}"}
\label{role-confusion}
}
\end{minipage}
\end{table*}

\subsection*{False Negative Spans}

False negatives in semantic role labeling are defined as the number of roles in the gold data that do not have the corresponding span matches in the model prediction. False negative for AGENT encompasses 69\% of errors from the total of 45 AGENT gold span role errors, while the errors in TIME roles all occur in this error type. In Table \ref{add-roles}, the left example shows that, the model failed to tag "\emph{ini komputer}" (EN: This is a computer). In the right example, the model did not recognize "\emph{get rick nya}\footnote{mistyped application name}" as PATIENT. An interesting remark is perhaps how the model failed to tag because the predicate is an unknown word in the training vocabulary despite the use of characters encoder to alleviate the out-of-vocabulary problem. While in the left example, predicate \emph{"menjawab"} is also an unknown word in the vocabulary but not a mistyped word, the right sample's predicate "\emph{di donlot}" is an informal writing of the word "download".

In the 50\% training data scenario, we found that multi-task active learning model achieves less recall compared to the single-task active learning model. The multi-task active learning with 50\% initial training data performance suffers from failing to tag 53\% of BENEFACTOR label.

\subsection*{Boundary Error}
Overall, we found that boundary errors contribute to 22\% of the total span exact match errors. For example, we found that PATIENT boundary errors mostly occurred because predicted role spans do not match the continuation of subsequent role. As shown in Table \ref{fix-boundary}, the model failed to recognize \emph{makanan} (EN: food) as the continuation of \emph{info} (EN: info) from the top example. In the bottom example, the model failed to predict the continuation of a mistyped role "\emph{sahabar}".

\subsection*{Role Confusion}
Role confusion is defined as the matching between gold span and predicted span, but they have different labels. This error typically occurs the least compared to the false negatives and boundary errors. In total, it is only 7\% of the total errors. The most common incorrect prediction is between gold PATIENT and prediction AGENT. As shown in Table \ref{role-confusion} in the top sentence, the model incorrectly labeled a PATIENT (Jemma) as an AGENT. Additionally, the model also incorrectly tagged BENEFACTOR as PATIENT. In the bottom sentence, the word "\emph{Aku}" (EN: I) is not annotated as any roles but detected as an AGENT by the model.

\section{Conclusion \& Future Work}
In this paper, we applied previous state-of-the-art deep semantic role labeling models on a low resource language in a conversational domain. We propose to combine multi-task and active learning methods into a single framework to achieve competitive SRL performance with less training data, and to leverage a semantically related task for SRL. 

Our primary motivation is to apply the framework for low resource languages in terms of dataset size and domains. Our experiments demonstrate that active learning method performs comparably well to the single-task baseline using 30\% fewer data by querying a total of 3483 from 4845 sentences. This result can be increased further marginally to outperform the baseline using 87\% of the training data. Our error analysis reveals some different obstacles from English SRL to work on in the future.

While \citeauthor{he2017deep}'s model of deep layers of highway LSTM allows learning the relation between a predicate and arguments explicitly, not all tasks in multi-task learning have equal complexity that needs deep layers. \citet{lower-level-mt} proposed a method to allow a model to predict tasks with different complexities at different layer depths. For example, predicting entity recognition tag at lower layers or inserting predicate features at higher layers in an LSTM, because entity recognition does not need predicates as features and is considered as a lower-level task compared to SRL. 

Combining multi-task learning with an unsupervised task such as language modeling \cite{Rei2017SemisupervisedML} is also a possible improvement in multi-task active learning settings as a semi-supervised variant. Analyzing other active learning methods such as query by committee, variance reduction \cite{Settles:2008:AAL:1613715.1613855}, and information density \cite{Wang2017ActiveLF} in multi-task settings are also a promising path in deep learning architectures. 

\bibliography{acl2018}
\bibliographystyle{acl_natbib}

\end{document}